%% file: main.tex
\documentclass[letterpaper, 10 pt, conference]{ieeeconf}

\IEEEoverridecommandlockouts
\usepackage{amsfonts}
\usepackage{amsmath}
\usepackage{amssymb}
\usepackage{graphicx}
\usepackage{siunitx}
\usepackage{pifont}
\usepackage{mathtools}

\usepackage{booktabs}
\usepackage{microtype}

\usepackage[dvipsnames]{xcolor}
\let\labelindent\relax
\usepackage[inline]{enumitem}
\usepackage{cite}
\usepackage{xspace}

\usepackage{flushend}
\renewcommand{\baselinestretch}{0.995}

\makeatletter
\let\NAT@parse\undefined
\makeatother

\usepackage[hang,flushmargin]{footmisc}
\usepackage[pdfencoding=auto, colorlinks=true]{hyperref}

\newcommand{\cmark}{\ding{51}}
\newcommand{\xmark}{\ding{55}}

\newcommand{\method}{\mbox{CounterPlay}\xspace}
\newcommand{\anchorbudget}{100B\xspace}
\newcommand{\postbudget}{1B\xspace}
\newcommand{\tabnote}[2]{\par\vspace{2pt}{\scriptsize\raggedright Table~\ref{#1}: #2\par}}
\title{\LARGE \bf
CounterPlay: Counterfactual Post-Training
for Self-Play Driving Policies}
\author{
Jiarong Wei$^{1,3}$,
Yin Wu$^{2,3}$,
Runkai He$^{3}$,
and Abhinav Valada$^{1}$
\thanks{$^1$ Department of Computer Science, University of Freiburg, Germany.
$^2$ Karlsruhe Institute of Technology, Germany.
$^3$ CARIAD SE, Germany. This work was partially funded by CARIAD SE.}
}
\hypersetup{
  pdftitle={CounterPlay: Counterfactual Post-Training for Self-Play Driving Policies},
  pdfauthor={Jiarong Wei, Yin Wu, Runkai He, Abhinav Valada}
}

\begin{document}
\maketitle
\thispagestyle{empty}
\pagestyle{empty}

\begin{abstract}
    \input{sections/0_abstract}
\end{abstract}

\input{sections/1_introduction}
\input{sections/2_related_work}
\input{sections/3_method}
\input{sections/4_experiment}
\input{sections/5_conclusion}

\bibliographystyle{IEEEtran}
\bibliography{reference}

\end{document}

%% file: sections/0_abstract.tex
Self-play in high-throughput simulators yields driving policies with robust closed-loop performance, but improvement per unit of simulation diminishes as training scales. Policies learn to handle common situations early, while further rollouts repeatedly encounter unresolved failures. Post-training offers an opportunity to target these failures, but existing methods primarily evaluate alternative actions or continuations at visited states, although successful recovery may require changing driving style earlier. We propose \method, a counterfactual self-play post-training approach that backtracks from failed tasks and retries them under alternate driving styles. \method rests on three key components. First, failure-driven backtracking uses the policy's value estimates to select an earlier stored state from which to retry the task. Second, reward conditioning enables a single policy to retry the task from this state using candidate styles ranging from cautious to aggressive. Third, \method retains task-completing retries only if no other vehicle incurs a new or earlier collision or off-road event relative to the factual branch. Retries that pass verification with fresh randomness are then distilled into the policy under its deployment condition. On BehaviorBench, \method achieves state-of-the-art scores on both the Interactive and Random splits across all eight traffic regimes using \postbudget post-training transitions, which is just 1\% of the anchor's \anchorbudget self-play training budget. Improvements over the anchor hold across all three evaluated driving styles. \method resolves a substantial fraction of the anchor's timeout cases on BehaviorBench and achieves a balance between task completion and safety that neither continued self-play nor adopting a more aggressive driving style attains.

%% file: sections/1_introduction.tex
\section{Introduction}

Self-play has emerged as an effective approach to learning driving policies, with a single policy controlling simulated vehicles and learning from billions of interactions with copies of
itself~\cite{cusumano2025gigaflow,cornelisse2025selfplay}. However, increasing the simulation budget yields diminishing returns. Frequently encountered situations are mastered early, while difficult cases persist despite repeated exposure. In these cases, additional rollouts can reproduce the same unsuccessful behavior, motivating a targeted approach to learn from the remaining failures.
Post-training provides a dedicated stage for addressing these residual failures and is increasingly used to improve driving policies~\cite{peng2024improving,yang2026posttraining}. Existing methods
primarily seek corrections at states visited by the policy. Group-relative and critic-based objectives score alternative actions or continuations from these states~\cite{kazemnejad2025vineppo,yu2025dapo,foerster2018coma}, while
counterfactual fine-tuning explores branches against replayed or modeled
traffic~\cite{li2026plannerrft,chen2026craft}. In self-play, however, any recorded state can be re-driven, so a lost task can be replayed from a point before the failure in a different driving style. This raises the central question of our work: can an earlier change of driving style resolve persistent self-play failures and provide useful post-training supervision?

\begin{figure}[t]
    \centering
    \includegraphics[width=\linewidth]{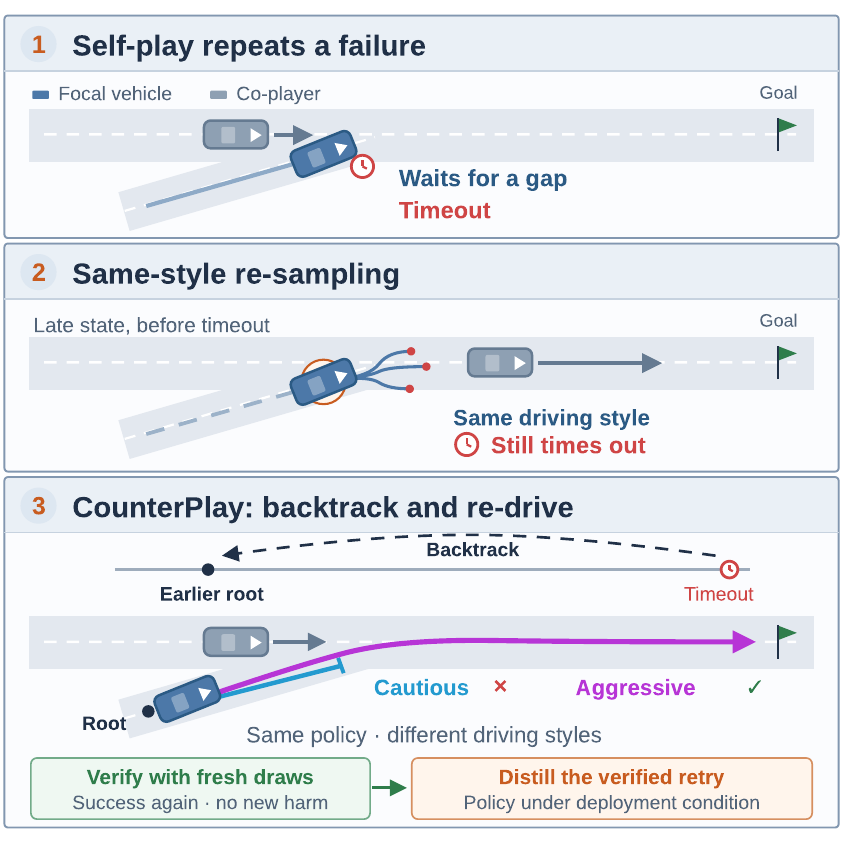}
    \caption{The self-play policy repeatedly waits for a gap and times out. Re-sampling the same driving style from a late state fails to recover the task. \method backtracks to an earlier root and explores alternative driving styles. An aggressive retry completes the task, passes verification with fresh randomness, and is distilled under the deployment condition.}
    \label{fig:cover}
    \vspace*{-0.4cm}
\end{figure}

Learning from these counterfactual retries presents three challenges.
The first is the intervention point. An intervention
placed after the decisive moment can no longer change the outcome,
whereas one placed long before it no longer isolates the decision under
study. The second is how to drive from that point. Most failures of a well-trained self-play policy are timeouts in which it keeps yielding or stays deadlocked without a timely decision, so a retry has to resolve them without giving up the safety the policy has learned. The third is counterfactual rollouts. A retry may score well under its own objective without completing the task, succeed by chance or at a neighbor's expense, and drive in a style the policy does not deploy and should not simply copy.

In this work, we propose \method, a counterfactual self-play post-training approach that backtracks from failed tasks, explores alternative driving styles, and distills verified recoveries into the deployment policy, as illustrated in Fig.~\ref{fig:cover}. Failure-driven backtracking uses the policy's recorded value estimates to choose an earlier state from which to retry the failed task. From this state, reward conditioning enables the same policy to execute candidate driving styles ranging from cautious to aggressive, providing temporally extended behavioral alternatives. We retain task-completing retries only if no other vehicle incurs a new or earlier collision or off-road event relative to the factual branch. We then verify their success with fresh randomness. These verified retries then supervise distillation under the policy's deployment condition. We evaluate \method on BehaviorBench~\cite{distelzweig2026behaviorbench} against post-training baselines with the same interaction budget and show that \method achieves state-of-the-art scores on both splits, and the gains hold across all evaluated driving styles. The approach resolves a substantial fraction of the anchor's timeout cases on BehaviorBench and achieves a balance between task completion and safety that neither continued self-play nor adopting a more aggressive driving style attains.

To summarize, our main contributions are as follows:
\begin{enumerate}
    \item We propose \method, a counterfactual self-play post-training
    approach that repairs failures self-play repeats by backtracking failed tasks and retrying them with alternate driving styles, turning verified recoveries into supervision for the deployment policy.
    \item We introduce failure-driven backtracking, which uses the policy's recorded value estimates to choose an earlier intervention point along a failed trajectory.
    \item We develop a reward-conditioned retry and distillation procedure that verifies task completion, checks for adverse effects on other vehicles, and tests success under fresh randomness before transferring corrective behavior to the deployment condition.
    \item We evaluate \method on \mbox{BehaviorBench} against post-training baselines with the same budget, achieving state-of-the-art scores on both splits across all the traffic regimes, while balancing task completion and safety.
\end{enumerate}

%% file: sections/2_related_work.tex
\section{Related Work}
\label{sec:related}

{
\noindent\textit{Self-Play for Autonomous Driving:}
In self-play, one policy typically controls every agent in the scene, so the training distribution evolves together with the learned behavior.
Nocturne~\cite{vinitsky2022nocturne} and GPUDrive~\cite{kazemkhani2025gpudrive} provide multi-agent driving simulation based on recorded scenarios for self-play training.
Gigaflow~\cite{cusumano2025gigaflow} shows that robust and naturalistic driving emerges from self-play alone over 1.6 billion kilometers of simulated experience.
Cornelisse~\textit{et al.}~\cite{cornelisse2025selfplay} scale self-play across thousands of recorded scenarios~\cite{ettinger2021womd} and achieve near-perfect goal completion on held-out scenes.
They also show that the pretrained policy can be fine-tuned within minutes on a few hand-designed rare scenarios.
CoPark~\cite{wei2026copark} applies self-play to reactive parking.
Recent studies identify limitations of self-play, including traffic-rule violations~\cite{sisask2026selfplay} and driving conventions incompatible with human behavior~\cite{cornelisse2026spiced}.
BehaviorBench~\cite{distelzweig2026behaviorbench} evaluates driving policies against eight frozen traffic-agent regimes, ranging from rule-based followers~\cite{treiber2000idm} to log replay.
To our knowledge, no prior work studies how a competent self-play policy can use a fixed additional budget to address recurring failures in its training scenarios.
}

{\parskip=2pt
\noindent\textit{Policy Post-Training:}
Post-training refines a pretrained policy in a separate stage with its own objective.
For language models, GRPO~\cite{shao2024grpo} and DAPO~\cite{yu2025dapo} compute advantages from groups of responses to the same prompt without a learned critic.
VinePPO~\cite{kazemnejad2025vineppo} instead estimates intermediate state values through Monte Carlo continuations from partial contexts.
For driving, Peng~\textit{et al.}~\cite{peng2024improving} fine-tune pretrained behavior models in a closed loop, and ADV-0~\cite{nie2026adv0} and AWM~\cite{nie2026awm} adapt a planner against traffic trained to expose its failures.
SPICED~\cite{cornelisse2026spiced} anchors self-play to a small human dataset through a KL term toward a behavior-cloning policy, and CL4AD~\cite{koprulu2026cl4ad} uses a curriculum to select self-play scenarios.
We focus on repairing a pretrained policy's failures by returning to earlier states and searching over its driving styles.
}

{\parskip=2pt
\noindent\textit{Counterfactual Learning for Driving Policies:}
Counterfactual methods differ in how they evaluate alternative decisions and where they intervene.
ParkDiffusion++~\cite{wei2026parkdiffusionpp} predicts the joint response of surrounding agents to alternative ego intentions with a learned model.
Another route executes the alternative.
The vine estimator of TRPO~\cite{schulman2015trpo} restores the simulator to visited states and rolls out several actions from each, and VinePPO~\cite{kazemnejad2025vineppo} revives this idea for credit assignment.
Go-Explore~\cite{ecoffet2021goexplore} and Backplay~\cite{resnick2018backplay} restore recorded states for exploration and curriculum learning, respectively.
For driving fine-tuning, PlannerRFT~\cite{li2026plannerrft} trains against log-replayed surrounding traffic.
CRAFT~\cite{chen2026craft} scores candidate trajectories with a hold-then-decay traffic proxy and supplements this signal with corrective advantages from executed closed-loop events.
Li and Kang~\cite{li2026datacentric} restore states near the end of failed trajectories, replace one action with an action from an expert or reference controller, and retain substitutions that reduce subsequent failure cost as supervision.
STaR~\cite{zelikman2022star} uses verified model-generated outputs as supervision.
Expert iteration~\cite{anthony2017exit} and policy distillation~\cite{rusu2016distillation} similarly train a network to imitate behavior produced by search or another policy.
However, these methods do not backtrack from a self-play policy's own failures to retry tasks under alternative ego reward conditions.
}

\begin{figure*}[t]
    \vspace*{4pt}
    \centering
    \includegraphics[width=\textwidth]{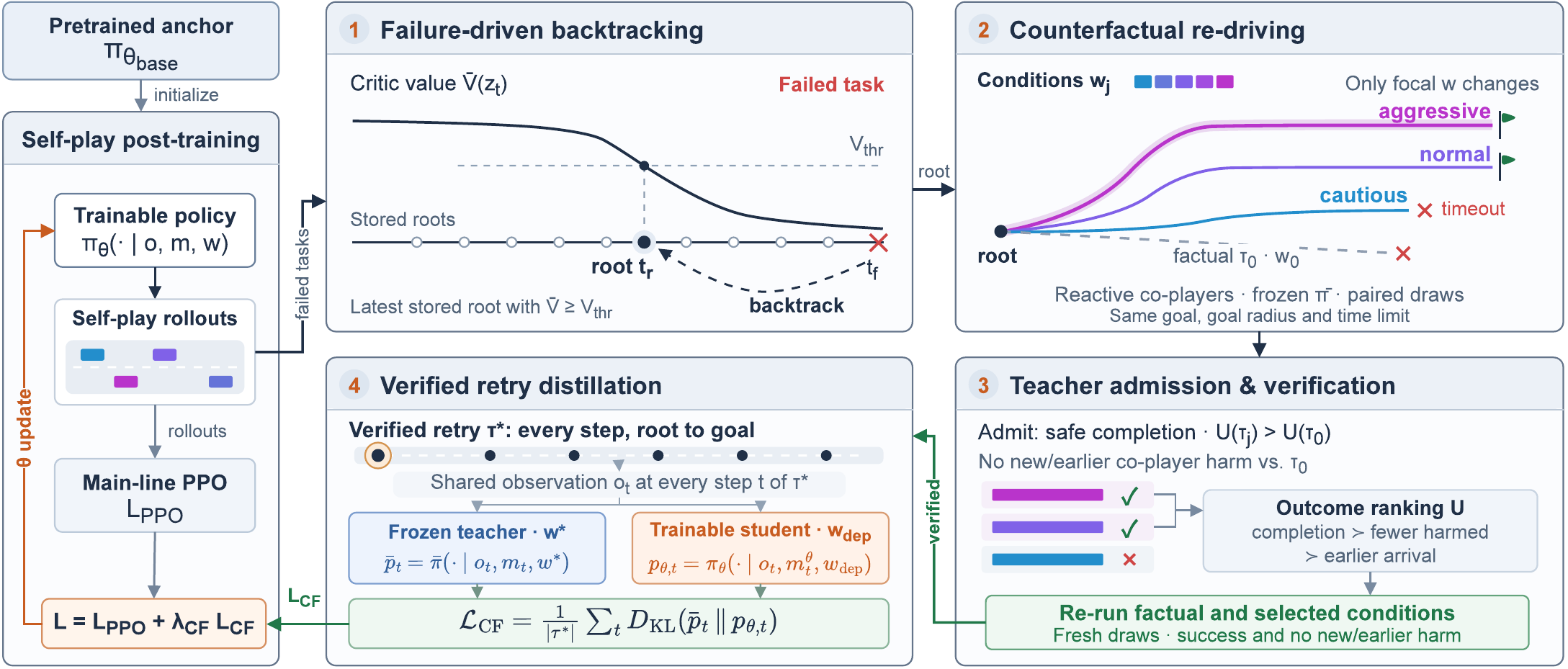}
    \vspace*{-0.5cm}
    \caption{\textbf{Overview of \method.} Based on a pretrained self-play anchor policy, Stage 1 uses critic estimates recorded along the failed trajectory to select an earlier root. Stage 2 then restores this state and simulates a factual branch under the original reward condition and retries under the candidate profiles. Stage 3 filters and ranks retries using task completion and other-vehicle outcomes relative to the factual branch. It then reruns the factual and selected branches with fresh shared random draws to verify the recovery. Stage 4 distills the verified retry into the policy under the deployment condition.}
    \label{fig:method_overview}
    \vspace*{-0.4cm}
\end{figure*}

{\parskip=2pt
\noindent\textit{Reward-Conditioned Policies and Behavior Control:}
Conditioning a policy on a reward specification turns one network into a family of behaviors.
Reward-conditioned policies~\cite{kumar2019rcp} learn actions as a function of a target return or advantage, Decision Transformer~\cite{chen2021dt} conditions a sequence model on the return-to-go, and Gigaflow~\cite{cusumano2025gigaflow} samples per-agent reward coefficients during self-play so that one policy drives cautiously or assertively on demand.
CtRL-Sim~\cite{rowe2024ctrlsim} controls the aggressiveness of simulated agents by tilting the distribution over per-component returns-to-go, BehaviorBench~\cite{distelzweig2026behaviorbench} evaluates planners against aggressive, normal, and cautious traffic produced by one conditioned policy, and AWM~\cite{nie2026awm} switches the role condition of traffic agents to attribute credit among adversaries.
To our knowledge, no prior work has varied the ego policy's own reward condition as a counterfactual intervention on tasks it previously failed.
}

%% file: sections/3_method.tex
\section{Method}
\label{sec:method}

In this section, we present the key components of \method, as
illustrated in Fig.~\ref{fig:method_overview}.
Sec.~\ref{subsec:formulation} formalizes the self-play post-training problem. Sec.~\ref{subsec:branching} to Sec.~\ref{subsec:evaluation} introduce failure-driven backtracking, re-driving the task in other styles of the same policy, and verification and distillation of the retries. Sec.~\ref{subsec:integration} explains how all stages share one post-training budget and enter the training loop.

\subsection{Problem Formulation}
\label{subsec:formulation}

We model driving as a partially observable Markov game
$\mathcal{G}=(\mathcal{N},\mathcal{S},\{\mathcal{A}^i\},\{\mathcal{O}^i\},P,\{r^i\},\gamma)$
among agents $i\in\mathcal{N}$. At time $t$, each policy-controlled
agent $i$ receives a local observation $o_t^i\in\mathcal{O}^i$ of the
state $s_t\in\mathcal{S}$ and selects an action $a_t^i\in\mathcal{A}^i$.
The state evolves as $s_{t+1}\sim P(\cdot\mid s_t,\mathbf a_t)$ under
the joint action $\mathbf a_t$, and agent $i$ receives the reward
$r^i(s_t,\mathbf a_t)$. An agent's task is to reach a goal $g$ within the episode length $T$. It fails on a collision, an off-road event, or a timeout.

The base policy $\pi_{\theta_{\mathrm{base}}}$ that we post-train is
trained by self-play with reward
conditioning~\cite{cusumano2025gigaflow}. All policy-controlled agents
share one policy $\pi_\theta$ acting from the observation history. At the start of each episode, agent $i$ samples a condition $w^i\sim p_{\mathrm{train}}(w)$ that specifies its preferences for safety, comfort, lane position, and target speed, together with the goal radius.
This condition determines the reward $r^i(s_t,\mathbf a_t;w^i)$ and is supplied to both the policy and critic, with actions sampled as $a_t^i\sim\pi_\theta(\cdot\mid o_{\le t}^i,w^i)$.
We use PPO~\cite{schulman2017ppo} to maximize each agent's expected discounted return
\begin{equation}
 J(\theta)=\mathbb{E}_{\pi_\theta}\!\left[\sum_{t\ge 0}\gamma^{t}\,
 r^i(s_t,\mathbf a_t;w^i)\right],
 \label{eq:selfplay_objective}
\end{equation}
where the expectation runs over self-play episodes. At deployment, the policy receives a single fixed condition $w_{\mathrm{dep}}$ that selects its driving style.

Starting from $\theta_{\mathrm{base}}$, we seek to improve the policy under a fixed budget of additional simulator transitions.
Ordinary self-play continues to optimize Eq.~\eqref{eq:selfplay_objective} under the original condition distribution $p_{\mathrm{train}}$.
Our primary target is performance under the fixed deployment condition $w_{\mathrm{dep}}$, and we also evaluate the other conditions in the policy family.

\subsection{Failure-Driven Backtracking}
\label{subsec:branching}

Most self-play episodes already succeed, but further rollouts often reproduce the remaining failures.
A failure may originate in an earlier decision, so we use the critic's recorded value estimates to choose where to restart the failed task.

\subsubsection{Failure Detection}
Within the branch budget, we search once per failed task.
Each search tests several styles from one earlier state for the agent that failed, called the focal agent.

\subsubsection{Root State Recording}
During each training generation, we collect self-play experience using frozen copies $\bar\pi$ and $\bar V$ of the current policy and critic.
We record roots at a fixed stride along the self-play rollouts. Each root $z_t$ stores the simulator state and every agent's recurrent memory and reward condition so retries can resume the recorded interaction.

\subsubsection{Value-Guided Root Selection}
We first identify stored states whose critic estimates meet or exceed a threshold, then choose the one closest in time to the failure.
The value $\bar V(z_t)$ is the focal agent's critic estimate at a stored root $z_t$.
Within a window of stored roots before the failure, let $V_{\max}$ be the highest estimate and $V_f$ the estimate at failure.
The threshold
\begin{equation}
 V_{\mathrm{thr}}=V_{\max}-\rho\,(V_{\max}-V_f),\qquad \rho\in[0,1],
 \label{eq:threshold}
\end{equation}
interpolates between these values, with $\rho$ controlling the interpolation.
The selected root time $t_r$ is therefore
\begin{equation}
 t_r=\max\bigl\{t:\ \bar V(z_t)\ge V_{\mathrm{thr}}\bigr\}.
 \label{eq:root}
\end{equation}
If no stored root in the window meets the threshold, we take its earliest root instead.
Critic values guide the starting point. The simulated outcome determines whether the retry succeeds.

\subsection{Reward-Conditioned Counterfactual Re-Driving}
\label{subsec:branches}

Reward conditioning lets the same policy drive in different styles, so we search for a recovery by changing the focal agent's condition and retrying from the selected root while keeping the policy frozen.

\subsubsection{Counterfactual Branch Generation}
For every branch, we restore the same simulator state and every agent's recorded recurrent memory from $z_{t_r}$.
Using the frozen policy $\bar\pi$, we simulate a factual branch $\tau_0$ under the focal agent's original condition $w_0$ and $J$ candidate retries $\tau_j$ under conditions $w_j\in\mathcal W$.
The other agents keep their own conditions and react to the evolving interaction in each branch.
Our fixed set $\mathcal W=\{w_1,\dots,w_J\}$ contains the benchmark's cautious, normal and aggressive profiles and the two midpoint profiles between adjacent styles.
We change only the style-related reward coefficients. The original goal, goal radius, arrival-speed threshold and deadline remain fixed.
Task-defining parameters are unchanged in both the policy input and the simulator's completion and termination checks.

\subsubsection{Shared Randomness Across Branches}
\label{subsubsec:sampling}
When retries use independent action samples, differences in their outcomes can reflect sampling randomness as well as driving style.
To control this source of randomness, we share random draws across branches from the same root.
At each time step $t$, we draw $u_t^n\sim\mathcal U(0,1)$ for each agent $n$ and use it to sample from that agent's action distribution in every branch:
\begin{equation}
 a^{n}_{t,j}=F^{-1}\bigl(u^{n}_{t};\,
 \bar\pi(\cdot\mid o^{n}_{t,j},m^{n}_{t,j},w^{n}_{j})\bigr),
 \label{eq:paired_sampling}
\end{equation}
where $F^{-1}(\cdot\,;p)$ is the inverse cumulative distribution function of the categorical action distribution $p$.
Here $m^n_{t,j}$ is agent $n$'s recurrent memory in branch $j$.
The condition $w_j^n$ equals $w_j$ for the focal agent. The other agents retain their own conditions.
Shared draws can still yield different actions because each branch has its own action distribution.

\subsubsection{Branch Termination and Co-Player Checking}
To compare effects on other vehicles over the same time horizon, we simulate each branch until the original deadline, even after the focal task ends.
We record the focal agent's outcome when its task ends and evaluate the other vehicles over the full branch.

\subsection{Teacher Verification and Distillation}
\label{subsec:evaluation}

We transfer recoveries found under alternative reward profiles into the deployment condition by selecting, verifying and distilling successful retries.

\subsubsection{Condition-Independent Branch Ranking}
To compare retries generated under different reward profiles, we rank them by the same task-outcome criteria rather than their profile-specific returns.
For a branch $\tau$, let $c(\tau)\in\{0,1\}$ indicate that the focal agent completes the task in time without a collision or off-road event, $n_{\mathrm{harm}}(\tau)$ the number of other vehicles that collide or go off-road along the branch, and $t_{\mathrm{arr}}(\tau)$ the arrival time ($+\infty$ when $c(\tau)=0$).
The resulting ranking is
\begin{equation}
 U(\tau)=\bigl(c(\tau),\;-n_{\mathrm{harm}}(\tau),\;-t_{\mathrm{arr}}(\tau)\bigr).
 \label{eq:utility}
\end{equation}
Completion takes priority. Among branches with the same completion outcome, we prefer fewer affected vehicles and then earlier arrival.

\subsubsection{Teacher Admission and Verification}
A retry is eligible to become a teacher only if it completes the original task without a collision or off-road event, $c(\tau_j)=1$, and outranks the factual branch, $U(\tau_j)>U(\tau_0)$.
It must also satisfy a constraint for every other vehicle: relative to the factual branch $\tau_0$, no collision or off-road event may be new or occur earlier.
We select the eligible retry trajectory $\tau^*$ ranked highest by the outcome ordering $U$.
For verification, we restore the root and rerun both factual and selected-condition branches once with fresh action-sampling draws shared between them.
The selected-condition rollout must again complete the original task without a collision or off-road event and meet the per-vehicle safety constraint relative to the new factual rollout.
If verification fails, we discard all retries from that search.

\subsubsection{Retry Distillation}
We distill each verified retry $\tau^*$ from the root until the focal task ends, training the policy under $w_{\mathrm{dep}}$ to imitate behavior generated under the selected retry's condition $w^*$.
At each recorded step, the frozen teacher uses the observation $o_t$ and recurrent memory $m_t$ under condition $w^*$ to produce $\bar p_t=\bar\pi(\cdot\mid o_t,m_t,w^*)$.
The student's recurrent memory is initialized from the memory stored at the root, without replaying observations from before that point.
It then processes the same observation sequence under $w_{\mathrm{dep}}$, updating its own recurrent memory $m_t^\theta$ to produce $p_{\theta,t}=\pi_\theta(\cdot\mid o_t,m_t^\theta,w_{\mathrm{dep}})$.
We average the KL divergence over each retry's steps and then over verified retries:
\begin{equation}
 \mathcal L_{\mathrm{CF}}(\theta)=
 \mathbb E_{\tau^{*}}\!\left[\frac{1}{|\tau^{*}|}\sum_{t}
 D_{\mathrm{KL}}\!\left(\bar p_t\,\big\|\,p_{\theta,t}\right)\right].
 \label{eq:cf_loss}
\end{equation}
The policy update minimizes the combined objective
\begin{equation}
 \mathcal L(\theta)=\mathcal L_{\mathrm{PPO}}(\theta)
 +\lambda_{\mathrm{CF}}\,\mathcal L_{\mathrm{CF}}(\theta),
 \label{eq:joint_objective}
\end{equation}
where $\mathcal L_{\mathrm{PPO}}$ is the negated clipped PPO surrogate for Eq.~\eqref{eq:selfplay_objective}, computed from the ordinary self-play rollout.
Branch trajectories contribute only through $\mathcal L_{\mathrm{CF}}$. They are not used for PPO or value-function updates.
If no verified teacher is available, the generation uses $\mathcal L_{\mathrm{PPO}}$ alone.

\begin{table*}[t]
\vspace*{6pt}
\centering
\caption{Benchmark scores per traffic regime on BehaviorBench.}
\vspace{-3pt}
\label{tab:main_results}
\scriptsize
\setlength\tabcolsep{1.6pt}
\begin{tabular*}{\textwidth}{@{\extracolsep{\fill}}lcccccccccccccccccc@{}}
\toprule
& \multicolumn{9}{c}{\textbf{Interactive1k}} & \multicolumn{9}{c}{\textbf{Random1k}} \\
\cmidrule(lr){2-10}\cmidrule(l){11-19}
\textbf{Planner} & IDM & PPO & SMART & Exp. & Aggr. & Norm. & Caut. & Mix & \textbf{Mean}
                 & IDM & PPO & SMART & Exp. & Aggr. & Norm. & Caut. & Mix & \textbf{Mean} \\
\midrule
Anchor & 48.95 & 66.84 & 59.00 & 55.68 & 68.48 & 60.64 & 49.39 & 59.29 & 58.53 & 62.64 & 69.11 & 63.97 & 62.78 & 71.79 & 70.02 & 66.05 & 68.03 & 66.80 \\
\midrule
PPO-Continue & \underline{50.92} & 65.27 & 56.82 & 53.53 & 69.47 & 61.93 & 51.00 & 60.58 & 58.69{\tiny$\pm$0.06} & 65.29 & 70.92 & 64.50 & 66.09 & 71.64 & 70.92 & 66.55 & 69.67 & 68.20{\tiny$\pm$0.13} \\
PPO-Rewind & 50.53 & 64.98 & 57.17 & 54.08 & 68.99 & 62.16 & 50.76 & 58.97 & 58.46{\tiny$\pm$0.21} & 65.63 & 71.12 & 65.03 & 65.54 & 71.77 & 70.86 & 66.83 & 69.35 & 68.27{\tiny$\pm$0.29} \\
PPO-Opponent & 49.90 & 66.33 & 57.33 & 53.91 & 69.37 & 63.04 & 50.42 & 59.11 & 58.68{\tiny$\pm$0.38} & 64.95 & 70.81 & 64.52 & \underline{66.34} & 71.82 & 70.91 & \underline{67.72} & 69.67 & 68.34{\tiny$\pm$0.13} \\
\midrule
SPICED$^\dagger$~\cite{cornelisse2026spiced} & 44.91 & 62.96 & \underline{59.83} & \underline{55.69} & 68.02 & \underline{65.35} & \underline{51.72} & 61.02 & 58.69{\tiny$\pm$0.94} & \underline{69.29} & \underline{73.78} & \underline{67.59} & 62.74 & \underline{75.17} & \underline{72.48} & 67.65 & \underline{70.40} & \underline{69.89}{\tiny$\pm$0.44} \\
CL4AD$^\dagger$~\cite{koprulu2026cl4ad} & 50.30 & 64.96 & 57.82 & 53.93 & \underline{70.25} & 61.84 & 50.94 & 60.26 & \underline{58.79}{\tiny$\pm$0.91} & 65.93 & 71.40 & 64.94 & 65.51 & 71.66 & 70.93 & 66.86 & 69.66 & 68.36{\tiny$\pm$0.16} \\
DAPO$^\dagger$~\cite{yu2025dapo} & 49.00 & 64.53 & 58.93 & 53.44 & 68.64 & 61.35 & 50.62 & 60.09 & 58.33{\tiny$\pm$0.35} & 66.20 & 69.33 & 64.42 & 63.92 & 72.74 & 70.86 & 67.14 & 68.47 & 67.88{\tiny$\pm$0.10} \\
VinePPO$^\dagger$~\cite{kazemnejad2025vineppo} & 50.18 & 64.16 & 57.19 & 53.04 & 69.46 & 61.57 & 50.49 & 60.41 & 58.31{\tiny$\pm$0.16} & 66.60 & 70.61 & 64.70 & 66.10 & 71.58 & 70.87 & 66.83 & 69.32 & 68.33{\tiny$\pm$0.13} \\
\midrule
PlannerRFT$^\dagger$~\cite{li2026plannerrft} & 48.86 & 64.68 & 59.34 & 53.24 & 68.91 & 62.46 & 50.13 & 60.91 & 58.57{\tiny$\pm$0.37} & 66.22 & 69.49 & 64.72 & 63.85 & 72.70 & 70.88 & 67.37 & 67.95 & 67.90{\tiny$\pm$0.12} \\
CRAFT$^\dagger$~\cite{chen2026craft} & 48.43 & \underline{66.55} & 59.58 & 52.25 & 69.67 & 61.27 & 49.53 & \underline{61.90} & 58.65{\tiny$\pm$0.84} & 64.10 & 70.08 & 63.69 & 63.76 & 72.70 & 70.87 & 67.09 & 67.94 & 67.53{\tiny$\pm$0.21} \\
\midrule
\method (Ours) & \textbf{59.73} & \textbf{73.53} & \textbf{65.53} & \textbf{63.51} & \textbf{76.44} & \textbf{70.22} & \textbf{56.65} & \textbf{67.30} & \textbf{66.61}{\tiny$\pm$0.14} & \textbf{75.04} & \textbf{78.82} & \textbf{71.83} & \textbf{70.21} & \textbf{81.01} & \textbf{79.55} & \textbf{75.44} & \textbf{75.59} & \textbf{75.94}{\tiny$\pm$0.70} \\
\bottomrule\end{tabular*}
\tabnote{tab:main_results}{We show the main results here. The Mean columns average eight regimes and also report standard deviations across three seeds. Best in bold, second best underlined among post-trained methods. The regimes comprise IDM, released PPO, SMART~\cite{wu2024smart}, expert log replay, three reward-conditioned PPO styles, and their per-agent mixture. Scores use the released BehaviorBench evaluator and are scaled by 100. On both 1,000-task splits, the ego selects argmax actions under the normal condition.}
\vspace*{-0.2cm}
\end{table*}

\subsection{Self-Play Training Integration}
\label{subsec:integration}

The post-training budget includes agent transitions from ordinary self-play and all branch simulations:
\begin{equation}
B_{\mathrm{post}}=B_{\mathrm{main}}+B_{\mathrm{factual}}+B_{\mathrm{candidate}}.
 \label{eq:post_budget}
\end{equation}
Here $B_{\mathrm{main}}$ counts ordinary self-play, $B_{\mathrm{factual}}$ counts factual branches from search and verification, and $B_{\mathrm{candidate}}$ counts candidate retries and verification of the selected condition.
All agent transitions count equally, including those from rejected retries.
Within each generation, branch transitions are capped at $\rho_{\mathrm{CF}} B_{\mathrm{main}}$. 
We run search branches in batches under the current generation's frozen policy alongside the next generation's self-play rollouts.
We then train on those rollouts with PPO and distill the verified recoveries in the same update.

%% file: sections/4_experiment.tex
\section{Experiment}
\label{sec:exp}

\begin{figure*}[t]
    \centering
    \includegraphics[width=\textwidth]{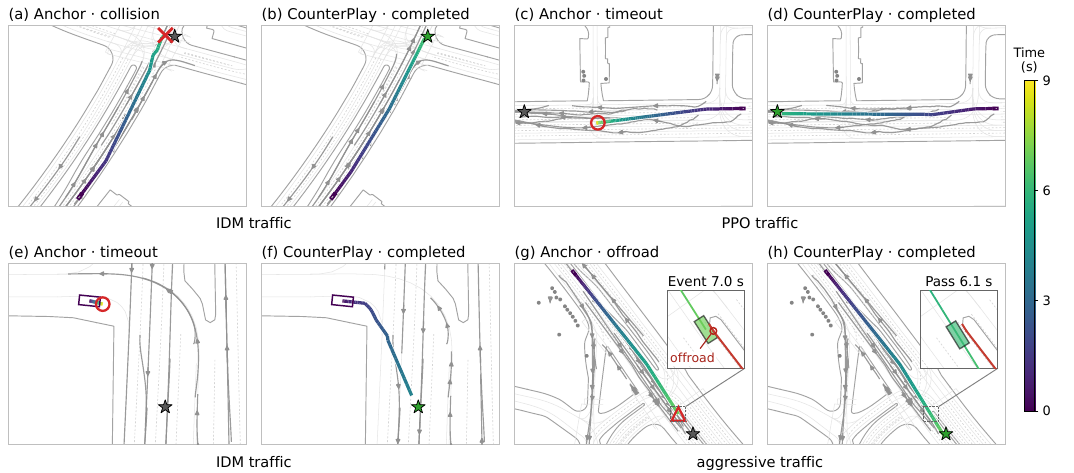}
    \vspace{-14pt}
    \caption{\textbf{Qualitative results.} Each pair compares the anchor (left) and \method (right) in the same scene and traffic regime. Ego trajectories are colored by elapsed time in seconds. Grey arrows show other road users' motion. Open boxes mark ego starts and stars mark goals. Red crosses mark collisions. Circles and triangles mark timeouts and off-road events, respectively.}
    \label{fig:qualitative}
    \vspace*{-0.35cm}
\end{figure*}

\subsection{Experimental Setup}
\label{subsec:setup}

\subsubsection{Benchmark and Data}
We evaluate on BehaviorBench~\cite{distelzweig2026behaviorbench}, a
closed-loop planning benchmark on the Waymo Open Motion
Dataset~\cite{ettinger2021womd}. The evaluated policy controls one designated ego vehicle.
Other agents follow one of eight frozen traffic regimes, from rule-based IDM~\cite{treiber2000idm}
to log replay (Tab.~\ref{tab:main_results}). \emph{Interactive1k} contains the 1,000 validation tasks with the highest interaction scores.
\emph{Random1k} contains 1,000 uniformly sampled validation tasks. Both splits are reserved for evaluation.
Post-training uses PufferDrive~\cite{cornelisse2025pufferdrive,suarez2025pufferlib} on the 80,000 WOMD training scenarios.
Episodes last 91 decisions with a decision interval of \SI{0.1}{\second}.
Each agent receives one logged goal. Agents that collide or go off-road are removed for the rest of the episode.

\subsubsection{Training Protocol}
Every method starts from an anchor, which is a reward-conditioned self-play PPO policy trained for \anchorbudget transitions following BehaviorBench's ~\cite{distelzweig2026behaviorbench} reward conditioning training recipe.
This network also generates the three conditioned traffic regimes and their per-agent mixture. Each method receives the same post-training
budget of \postbudget agent transitions, counted as $B_{\mathrm{post}}$ defined in
Eq.~\eqref{eq:post_budget}. We post-train each method with three training seeds under the anchor's condition distribution $p_{\mathrm{train}}$.
Evaluation uses a fixed deployment condition $w_{\mathrm{dep}}$. The main comparison uses normal deployment. Tab.~\ref{tab:deployment} compares the other styles.

\subsubsection{Baselines}
\emph{PPO variants.} \emph{PPO-Continue} spends the budget on factual
self-play. \emph{PPO-Rewind} uses the same failure criteria as \method and restores the state eight decisions before the focal agent's failure.
Ordinary PPO continues from that state under the original condition.
\emph{PPO-Opponent} drives a fraction of the agents with earlier policy checkpoints.

\emph{Recent post-training methods.} We adapt these methods to the same recurrent policy and simulator. A dagger marks each adaptation.
\emph{SPICED}$^\dagger$~\cite{cornelisse2026spiced} adds a human-anchoring KL penalty to PPO.
Its reference is a frozen behavior-cloning policy trained on log-tracking actions from the training scenarios.
\emph{CL4AD}$^\dagger$~\cite{koprulu2026cl4ad} replaces uniform scenario sampling with a learnability curriculum.
It emphasizes scenarios that the policy sometimes completes and sometimes fails.
\emph{DAPO}$^\dagger$~\cite{yu2025dapo} and
\emph{VinePPO}$^\dagger$~\cite{kazemnejad2025vineppo} restore states where the ego approaches another vehicle. Both execute groups of closed-loop continuations.
DAPO computes group-relative advantages and retains its decoupled clipping and dynamic sampling.
VinePPO replaces the critic's root-value estimate with a Monte Carlo estimate from these continuations.

\emph{Counterfactual post-training.}
\emph{PlannerRFT}$^\dagger$~\cite{li2026plannerrft} uses a clipped group-relative trajectory objective and evaluates continuations against log-replayed vehicles.
Its original PPO-trained exploration policy controls a diffusion denoiser's guidance scale.
Our categorical policy has no denoiser, so a learned exploration head instead controls the sampling temperature of continuations.
PPO trains this head on the group return. The planner retains the clipped group-relative GRPO loss.
\emph{CRAFT}$^\dagger$~\cite{chen2026craft} uses a score-function objective with group-normalized counterfactual advantages.
We adapt its candidate-based formulation to continuations from the recurrent categorical policy.
Its traffic proxy holds other vehicles' last actions and then decays them.
We retain its residual correction from closed-loop rollouts under frozen co-players.
We also retain its asymmetric KL regularization toward an exponential moving average of the policy.
Both adaptations preserve the original methods' traffic assumptions.
All four continuation-based baselines match group sizes and branch horizons.
Their transitions count toward $B_{\mathrm{post}}$ at the same rate as branches of \method.

\subsubsection{Evaluation Metrics}
A task receives zero score if the ego misses the goal or incurs an at-fault collision or off-road event.
Otherwise, its score is
\begin{equation}
 S=0.2\,S_{\mathrm{cmf}}+0.5\,S_{\mathrm{align}}+0.3\,S_{\mathrm{ctr}},
\label{eq:benchmark_score}
\end{equation}
where $S_{\mathrm{cmf}}$, $S_{\mathrm{align}}$ and $S_{\mathrm{ctr}}$ measure comfort, lane alignment and lane centering, respectively.
Each lies in $[0,1]$.
Metric definitions and parameter values follow BehaviorBench~\cite{distelzweig2026behaviorbench}.

\subsubsection{Implementation Details}
All methods use the anchor's policy network and a learning rate of $3\times10^{-5}$ with a cosine schedule.
\method stores a root every five decisions. Value-guided selection uses $\rho=0.5$ within a window of 5--60 decisions before failure.
Each root runs one factual branch and $J=5$ candidate branches.
Branch transitions are capped at $\rho_{\mathrm{CF}}=50\%$ of main-line transitions. Their observed share of the total budget $B_{\mathrm{post}}$ is $27\%$.
We average the distillation KL over each teacher's steps and then over teachers in the generation.
The loss has weight $\lambda_{\mathrm{CF}}=0.5$ and contributes to four minibatch updates per generation.
Across training, $41\%$ of searched roots yield an admissible retry. Of these selected retries, $88\%$ pass verification, producing about 70 teachers per generation.

\subsection{Main Results}
\label{subsec:main_results}

\subsubsection{Quantitative Results}
Tab.~\ref{tab:main_results} compares methods after \postbudget post-training transitions using BehaviorBench's normal ego reward profile~\cite{distelzweig2026behaviorbench}.
Its moderate reward settings provide a common reference for comparing post-training gains.
\method leads on both splits in all eight traffic regimes, with mean gains over PPO-Continue of $7.9$ points on Interactive1k and $7.7$ on Random1k.
On Interactive1k, the gains are largest against log replay ($10.0$ points), IDM ($8.8$) and SMART ($8.7$).
They are smaller against the conditioned policy family, reaching a minimum of $5.7$ points against cautious traffic.
The larger gains against traffic outside this family show that \method responds more robustly to unfamiliar agents.
Even with reward conditioning, self-play exposes the anchor only to copies of its own policy.
\method, however, achieves broader robustness by learning from backtracked failures and distilling alternative driving styles.
The baselines remain within half a point of the anchor on Interactive1k despite the additional training budget.
These updates lack an explicit search for a different driving style at an earlier state.
Their limited gains suggest that continued training or changes to experience collection and the learning objective alone are insufficient to resolve persistent yielding and timeout failures after \anchorbudget self-play transitions.

\subsubsection{Qualitative Results}
Fig.~\ref{fig:qualitative} shows how \method resolves collision, timeout, and off-road failures encountered by the anchor.
At the intersection in (a,b), \method completes the crossing under IDM traffic while the anchor collides.
Both timeout cases involve prolonged hesitation: the anchor nearly stops under PPO traffic (c,d) and barely moves before a right turn under IDM traffic (e,f).
\method overcomes this stalled behavior to reach the goal in both scenes.
In the final pair (g,h), \method stays on-road and reaches the goal under aggressive traffic while the anchor goes off-road.

\subsection{Ablation Study}
\label{subsec:ablations}

\subsubsection{Component Ablations}
We examine the contributions of backtracking, reward conditioning, teacher selection, verification and distillation through the variants in Tab.~\ref{tab:buildup}.
For backtracking, we restore states 10, 30 or 50 decisions before failure instead of letting the critic choose the intervention point.
For search without a style change, retries retain the failed attempt's reward profile. The focal-agent action draws are independent across factual and candidate branches, while draws for the other agents remain shared.
In a separate test of teacher selection, we score each retry under the profile that generated it and select the highest-scoring retry instead of ranking candidates by task completion and safety.
We test verification by omitting the second execution used to check whether a selected retry succeeds again with fresh randomness.
We compare distillation with PPO on the same accepted recovery trajectories, using rewards for PPO updates instead of matching the teacher's action probabilities.

All fixed depths reduce performance.
Shallow roots may leave little room to change an interaction, while deeper roots may include more steps unrelated to its failure.
Under the evaluated sampling protocols, retries without a style change barely improve over PPO-Continue and remain $6.8$ points below \method.
Selection by reward also performs worse than selection using common task criteria.
Omitting verification changes the score little but raises the observed at-fault collision rate from $2.57\%$ to $3.1\%$ on Interactive1k.
This supports verifying recoveries before using them to train the policy.
PPO on the same recoveries also performs worse than learning directly from the teacher.

\begin{table}[!t]
\vspace*{6pt}
\centering
\caption{Component ablations on Interactive1k.}
\vspace{-3pt}
\label{tab:buildup}
\scriptsize
\setlength\tabcolsep{3pt}
\begin{tabular*}{\columnwidth}{@{\extracolsep{\fill}}lccccc@{}}
\toprule
\textbf{Variant} & \textbf{B} & \textbf{C} & \textbf{S} & \textbf{D} & \textbf{Score}$\uparrow$ \\
\midrule
Anchor (no post-training)  & -- & -- & -- & -- & 58.53 \\
PPO-Continue               & \xmark & \xmark & \xmark & \xmark & 58.69{\tiny$\pm$0.06} \\
\midrule
\multicolumn{6}{@{}l}{\emph{Backtracking}} \\
10 decisions before failure  & 10 & \cmark & \cmark & \cmark & 62.83{\tiny$\pm$0.41} \\
30 decisions before failure  & 30 & \cmark & \cmark & \cmark & 65.21{\tiny$\pm$0.33} \\
50 decisions before failure  & 50 & \cmark & \cmark & \cmark & 64.58{\tiny$\pm$0.39} \\
\midrule
\multicolumn{6}{@{}l}{\emph{Reward conditioning}} \\
No change in driving style             & \cmark & \xmark & \cmark & \cmark & 59.84{\tiny$\pm$0.27} \\
\midrule
\multicolumn{6}{@{}l}{\emph{Selection}} \\
Selection by reward         & \cmark & \cmark & \xmark & \cmark & 63.42{\tiny$\pm$0.46} \\
Without verification            & \cmark & \cmark & \cmark$^{-}$ & \cmark & \underline{66.20}{\tiny$\pm$0.31} \\
\midrule
\multicolumn{6}{@{}l}{\emph{Distillation}} \\
PPO on recovery trajectories                 & \cmark & \cmark & \cmark & \xmark & 61.40{\tiny$\pm$0.31} \\
\midrule
\method (full)             & \cmark & \cmark & \cmark & \cmark & \textbf{66.61}{\tiny$\pm$0.14} \\
\bottomrule
\end{tabular*}
\tabnote{tab:buildup}{For post-trained policies, we report the mean and standard deviation across three training seeds. Best in bold, second best underlined. B: backtracking (\cmark: value-guided, numbers: decisions before failure). C: search across reward profiles. S: task-based selection and verification (\cmark$^{-}$: no re-execution). D: teacher distillation. Without a style change, retries use $w_j=w_0$.}
\end{table}

\begin{table}[!t]
\centering
\caption{Task retention and failure repair after post-training (\%).}
\vspace{-3pt}
\label{tab:failures}
\scriptsize
\setlength\tabcolsep{3pt}
\begin{tabular*}{\columnwidth}{@{\extracolsep{\fill}}lccccc@{}}
\toprule
\textbf{Policy} & \textbf{Completed} & \textbf{Timeout} & \textbf{AF coll.} & \textbf{Non-AF coll.} & \textbf{Offroad} \\
\midrule
\multicolumn{6}{@{}l}{\emph{Interactive1k}} \\
Anchor   & 68.4 & 23.1 & 3.8 & 4.5 & 0.2 \\
\cmidrule(lr){1-6}
PPO-Continue     & \textbf{96.3}{\tiny$\pm$0.11} & 4.9{\tiny$\pm$0.5} & 30.0{\tiny$\pm$3.3} & 9.3{\tiny$\pm$1.6} & 33.3{\tiny$\pm$11.1} \\
\method          & 95.1{\tiny$\pm$0.59} & \textbf{45.9}{\tiny$\pm$1.4} & \textbf{38.9}{\tiny$\pm$5.1} & \textbf{38.0}{\tiny$\pm$2.6} & \textbf{40.7}{\tiny$\pm$12.8} \\
\midrule
\multicolumn{6}{@{}l}{\emph{Random1k}} \\
Anchor   & 75.6 & 18.5 & 1.9 & 1.8 & 2.2 \\
\cmidrule(lr){1-6}
PPO-Continue     & \textbf{99.2}{\tiny$\pm$0.17} & 6.8{\tiny$\pm$0.7} & 37.8{\tiny$\pm$3.9} & 12.9{\tiny$\pm$2.9} & 13.5{\tiny$\pm$2.6} \\
\method          & 97.2{\tiny$\pm$0.91} & \textbf{67.3}{\tiny$\pm$1.2} & \textbf{46.7}{\tiny$\pm$6.7} & \textbf{67.9}{\tiny$\pm$4.0} & \textbf{23.8}{\tiny$\pm$3.2} \\
\bottomrule
\end{tabular*}
\tabnote{tab:failures}{Anchor rows report the outcome distribution over $8{,}000$ task--regime evaluations per split. Post-trained rows report the percentage of each anchor class reaching the goal without a collision or off-road event. We report the mean and standard deviation across three training seeds. Bold marks the higher post-trained rate. Anchor classes prioritize at-fault (AF) collision, off-road, goal completion, non-AF collision, then timeout. Class shares are rounded independently.}
\vspace*{-0.2cm}
\end{table}

\subsubsection{Failure Repair Analysis}
We examine which failures account for \method's gains over continued self-play (Tab.~\ref{tab:failures}).
For each of Interactive1k and Random1k, we evaluate 1,000 tasks under all eight traffic regimes and group the results by the anchor's outcome.
We then compare PPO-Continue and \method within each group to identify which failures they repair and whether they retain success on previously completed tasks.
A repair requires the ego to reach its goal without a collision or off-road event. 
Repairing timeouts, the anchor's main failure mode, explains most of the gain in collision- and off-road-free completion.
On Interactive1k, \method repairs $45.9\%$ of these cases compared with $4.9\%$ for PPO-Continue. Random1k shows the same pattern.
Compared with PPO-Continue, \method completes slightly fewer tasks that the anchor had already solved.
It completes more tasks overall by repairing far more of the anchor's timeouts.
It also repairs more at-fault and non-at-fault collision cases.

\subsubsection{Deployment Condition Analysis}
\label{subsubsec:deployment}
The timeout results raise a natural question: could a more aggressive anchor achieve the same gains?
We compare the anchor, PPO-Continue and \method under the same ego deployment condition to separate post-training gains from changes in driving style (Tab.~\ref{tab:deployment}).
We also train two distillation controls under the normal reward input to imitate a frozen teacher's aggressive or cautious behavior, without failure backtracking, search or verification. 
We evaluate all five policies under normal and aggressive reward settings and the halfway point between them without further updating their weights.
\method achieves the highest score under all three deployment conditions and the lowest at-fault collision rate under halfway and aggressive deployment.
Under normal deployment, direct aggressive distillation achieves much of \method's score improvement over PPO-Continue.
However, \method achieves a higher score and a lower at-fault collision rate than direct aggressive distillation under each of the three deployment conditions.
Under normal deployment, cautious distillation has a lower at-fault collision rate than \method but a substantially lower score.
Increasing the anchor's deployment aggression improves its score but also increases at-fault collisions.
For \method itself, halfway deployment nearly matches the aggressive score with a lower observed at-fault collision rate ($1.79\%$ versus $1.92\%$), offering a favorable balance between performance and risk.

\begin{table}[!t]
\vspace*{6pt}
\centering
\caption{Effect of ego deployment condition on Interactive1k.}
\vspace{-3pt}
\label{tab:deployment}
\scriptsize
\setlength\tabcolsep{1pt}
\begin{tabular*}{\columnwidth}{@{\extracolsep{\fill}}lcccccc@{}}
\toprule
& \multicolumn{2}{c}{\textbf{Normal}} & \multicolumn{2}{c}{\textbf{Halfway}} & \multicolumn{2}{c}{\textbf{Aggressive}} \\
\cmidrule(lr){2-3}\cmidrule(lr){4-5}\cmidrule(lr){6-7}
\textbf{Method} & \textbf{Score}$\uparrow$ & \textbf{AF-C.}$\downarrow$ & \textbf{Score}$\uparrow$ & \textbf{AF-C.}$\downarrow$ & \textbf{Score}$\uparrow$ & \textbf{AF-C.}$\downarrow$ \\
\midrule
Anchor                   & 58.53 & 3.75 & 64.21 & 4.03 & 69.85 & 4.38 \\
PPO-Continue             & 58.69{\tiny$\pm$0.06} & \underline{2.17}{\tiny$\pm$0.14} & 64.73{\tiny$\pm$0.72} & 2.75{\tiny$\pm$0.50} & \underline{69.49}{\tiny$\pm$0.89} & 3.12{\tiny$\pm$0.62} \\
Aggressive distillation  & \underline{65.44}{\tiny$\pm$0.42} & 4.46{\tiny$\pm$0.35} & \underline{70.07}{\tiny$\pm$0.39} & 4.62{\tiny$\pm$0.33} & 66.22{\tiny$\pm$0.47} & 5.08{\tiny$\pm$0.41} \\
Cautious distillation    & 61.18{\tiny$\pm$0.48} & \textbf{1.86}{\tiny$\pm$0.22} & 65.83{\tiny$\pm$0.45} & \underline{1.90}{\tiny$\pm$0.24} & 68.94{\tiny$\pm$0.52} & \underline{2.20}{\tiny$\pm$0.27} \\
\method                  & \textbf{66.61}{\tiny$\pm$0.14} & 2.57{\tiny$\pm$0.29} & \textbf{73.55}{\tiny$\pm$0.18} & \textbf{1.79}{\tiny$\pm$0.07} & \textbf{73.60}{\tiny$\pm$0.75} & \textbf{1.92}{\tiny$\pm$0.51} \\
\bottomrule
\end{tabular*}
\tabnote{tab:deployment}{For post-trained policies, we report the mean and standard deviation across three training seeds. AF-C.: at-fault collision rate (\%). Best post-trained values in bold, second best underlined. Halfway averages the normal and aggressive reward settings. All conditions use the same goal criterion. Both distillation controls match \method's PPO update and budget.}
\end{table}

\begin{table}[!t]
\centering
\caption{Search with one or five driving styles on Interactive1k.}
\vspace{-3pt}
\label{tab:conditioning}
\scriptsize
\setlength\tabcolsep{3pt}
\begin{tabular*}{\columnwidth}{@{\extracolsep{\fill}}lcccc@{}}
\toprule
\textbf{Search styles} & \textbf{Score}$\uparrow$
& \textbf{AF-C.}$\downarrow$ & \textbf{Off.}$\downarrow$
& \textbf{Goal}$\uparrow$ \\
\midrule
All five styles (\method)     & \textbf{66.61}{\tiny$\pm$0.14} & \textbf{2.57}{\tiny$\pm$0.29} & \underline{0.67}{\tiny$\pm$0.07} & \textbf{79.0}{\tiny$\pm$0.14} \\
Cautious            & 57.02{\tiny$\pm$0.55} & 2.88{\tiny$\pm$0.31} & \textbf{0.50}{\tiny$\pm$0.12} & 66.5{\tiny$\pm$0.62} \\
Cautious--normal midpoint    & 54.27{\tiny$\pm$0.47} & \underline{2.75}{\tiny$\pm$0.26} & 1.62{\tiny$\pm$0.21} & 63.5{\tiny$\pm$0.53} \\
Normal              & 54.18{\tiny$\pm$0.52} & 3.58{\tiny$\pm$0.33} & 1.25{\tiny$\pm$0.18} & 63.5{\tiny$\pm$0.58} \\
Normal--aggressive midpoint  & 64.56{\tiny$\pm$0.37} & 3.84{\tiny$\pm$0.33} & 1.12{\tiny$\pm$0.15} & 75.3{\tiny$\pm$0.44} \\
Aggressive          & \underline{65.75}{\tiny$\pm$0.28} & 3.06{\tiny$\pm$0.27} & 0.88{\tiny$\pm$0.11} & \underline{77.9}{\tiny$\pm$0.35} \\
\bottomrule
\end{tabular*}
\tabnote{tab:conditioning}{We report the mean and standard deviation across three training seeds. Each search uses five candidate retries. Best in bold, second best underlined. Evaluation uses the normal ego condition. The normal-profile variant uses same-condition self-distillation and does not directly reinforce the sampled actions. AF-C., Off. and Goal are at-fault collision, off-road and completion rates (\%) over all eight regimes.}
\vspace*{-0.2cm}
\end{table}

\subsubsection{Driving Style Diversity}
The fixed-style controls above omit failure recovery search, so they do not establish whether searching with a single driving style would suffice.
We test this by keeping backtracking, teacher selection, verification and distillation unchanged while restricting every retry to one fixed reward profile (Tab.~\ref{tab:conditioning}).
During these searches, focal-agent action draws are independent across factual and candidate branches, while co-player draws remain shared.
Using only the cautious profile yields fewer at-fault collisions and off-road events than using only the aggressive profile, but also fewer completions.
Their complementary strengths motivate a search that provides both cautious responses to reduce risk and aggressive responses to make progress.
The full panel achieves higher completion and fewer at-fault collisions than every single-profile variant tested.

%% file: sections/5_conclusion.tex
\section{Conclusion}
\label{sec:conclusion}

We presented \method, a counterfactual post-training method that backtracks the failed tasks of a self-play policy and re-drives them in other styles of the same reward-conditioned policy. On BehaviorBench, it turns a large share of the anchor's timeouts into completions with a small post-training budget, while maintaining a balance between completion and safety that a fixed shift of the same policy to a more aggressive driving style does not achieve.

The main limitation lies in the source of the counterfactuals. The
retries are drawn from a fixed panel of five hand-chosen reward profiles, so the search can only find repair styles already in the panel, and the method does not learn to propose conditions tailored to each failure. A natural next step is to learn the condition space, or a style generator, so the search proposes more human-like, safer, and more comfortable driving than any preset profile. Measuring the outcomes of the other vehicles around the post-trained policy would test whether the teacher filter's safety benefit carries over to deployment, and a second anchor would test whether the post-training generalizes to policies from other self-play recipes and simulators.

%% file: reference.bib
@String { CVPR          = {{IEEE/CVF} Conf. on Computer Vision and Pattern Recognition} }

@String { ICRA          = {{IEEE} Int. Conf. on Robotics and Automation} }

@String { ICCV          = {{IEEE/CVF} Int. Conf. on Computer Vision} }

@String { ECCV          = {Eur. Conf. on Computer Vision} }

@String { NEURIPS       = {Conf. on Neural Information Processing Systems} }

@String { ICLR          = {Int. Conf. on Learning Representations} }

@String { ICML          = {Int. Conf. on Machine Learning} }

@String { CORL          = {Conf. on Robot Learning} }

@String { CAI           = {{AAAI} Conf. on Artificial Intelligence} }

@inproceedings{cusumano2025gigaflow,
  title={Robust Autonomy Emerges from Self-Play},
  author={Cusumano-Towner, Marco and Hafner, David and Hertzberg, Alexander and Huval, Brody and others},
  booktitle=ICML,
  pages={11710--11737},
  year={2025}
}

@article{cornelisse2025selfplay,
  title={Building Reliable Sim Driving Agents by Scaling Self-Play},
  author={Cornelisse, Daphne and Pandya, Aarav and Joseph, Kevin and Su{\'a}rez, Joseph and Vinitsky, Eugene},
  journal={arXiv preprint arXiv:2502.14706},
  year={2025}
}

@article{distelzweig2026behaviorbench,
  title={Beyond Self-Play and Scale: A Behavior Benchmark for Generalization in Autonomous Driving},
  author={Distelzweig, Aron and Janjo{\v{s}}, Faris and Look, Andreas and Rothenh{\"a}usler, Anna and others},
  journal={arXiv preprint arXiv:2605.10034},
  year={2026}
}

@article{cornelisse2026spiced,
  title={Human-like Autonomy Emerges from Self-Play and a Pinch of Human Data},
  author={Cornelisse, Daphne and Hunt, Julian and Zhang, Zixu and Doulazmi, Wa{\"e}l and Joseph, Kevin and {Fern{\'a}ndez Fisac}, Jaime and Vinitsky, Eugene},
  journal={arXiv preprint arXiv:2606.19370},
  year={2026}
}

@article{chen2026craft,
  title={{CRAFT}: Counterfactual-to-Interactive Reinforcement Fine-Tuning for Driving Policies},
  author={Chen, Keyu and Ye, Nanfei and Wang, Yida and Sun, Wenchao and Zhao, Danqi and Cheng, Hao and Zheng, Sifa},
  journal={arXiv preprint arXiv:2605.04470},
  year={2026}
}

@inproceedings{foerster2018coma,
  title={Counterfactual Multi-Agent Policy Gradients},
  author={Foerster, Jakob N. and Farquhar, Gregory and Afouras, Triantafyllos and Nardelli, Nantas and Whiteson, Shimon},
  booktitle=CAI,
  volume={32},
  pages={2974--2982},
  year={2018}
}

@inproceedings{kazemkhani2025gpudrive,
  title={{GPUDrive}: Data-Driven, Multi-Agent Driving Simulation at 1 Million {FPS}},
  author={Kazemkhani, Saman and Pandya, Aarav and Cornelisse, Daphne and Shacklett, Brennan and Vinitsky, Eugene},
  booktitle=ICLR,
  year={2025}
}

@article{suarez2025pufferlib,
  title={{PufferLib} 2.0: Reinforcement Learning at 1{M} steps/s},
  author={Suarez, Joseph},
  journal={Reinforcement Learning Journal},
  volume={6},
  pages={1378--1388},
  year={2025}
}

@misc{cornelisse2025pufferdrive,
  title={{PufferDrive}: A Fast and Friendly Driving Simulator for Training and Evaluating {RL} Agents},
  author={Cornelisse, Daphne and Cheng, Spencer and Mandavilli, Pragnay and Hunt, Julian and Joseph, Kevin and Doulazmi, Wa{\"e}l and Gupta, Aditya and Vinitsky, Eugene},
  howpublished={\url{https://github.com/Emerge-Lab/PufferDrive}},
  note={Version 2.0.0},
  year={2025}
}

@article{shao2024grpo,
  title={{DeepSeekMath}: Pushing the Limits of Mathematical Reasoning in Open Language Models},
  author={Shao, Zhihong and Wang, Peiyi and Zhu, Qihao and Xu, Runxin and Song, Junxiao and Bi, Xiao and Zhang, Haowei and Zhang, Mingchuan and Li, Y. K. and Wu, Y. and Guo, Daya},
  journal={arXiv preprint arXiv:2402.03300},
  year={2024}
}

@inproceedings{schulman2015trpo,
  title={Trust Region Policy Optimization},
  author={Schulman, John and Levine, Sergey and Abbeel, Pieter and Jordan, Michael and Moritz, Philipp},
  booktitle=ICML,
  pages={1889--1897},
  year={2015}
}

@inproceedings{kazemnejad2025vineppo,
  title={{VinePPO}: Refining Credit Assignment in {RL} Training of {LLMs}},
  author={Kazemnejad, Amirhossein and Aghajohari, Milad and Portelance, Eva and Sordoni, Alessandro and Reddy, Siva and Courville, Aaron and Le Roux, Nicolas},
  booktitle=ICML,
  pages={29557--29590},
  year={2025}
}

@inproceedings{li2026plannerrft,
  title={{PlannerRFT}: Reinforcing Diffusion Planners through Closed-Loop and Sample-Efficient Fine-Tuning},
  author={Li, Hongchen and Li, Tianyu and Yang, Jiazhi and Shang, Mingyang and Wu, Gaoqiang and Wang, Caojun and Tian, Haochen and Lin, Zengrong and Hao, Zhihui and Lang, XianPeng and Hu, Jia and Li, Hongyang},
  booktitle=CVPR,
  pages={24929--24938},
  year={2026}
}

@inproceedings{vinitsky2022nocturne,
  title={Nocturne: A Scalable Driving Benchmark for Bringing Multi-Agent Learning One Step Closer to the Real World},
  author={Vinitsky, Eugene and Lichtl{\'e}, Nathan and Yang, Xiaomeng and Amos, Brandon and Foerster, Jakob},
  booktitle=NEURIPS # { Datasets and Benchmarks Track},
  year={2022}
}

@article{schulman2017ppo,
  title={Proximal Policy Optimization Algorithms},
  author={Schulman, John and Wolski, Filip and Dhariwal, Prafulla and Radford, Alec and Klimov, Oleg},
  journal={arXiv preprint arXiv:1707.06347},
  year={2017}
}

@inproceedings{ettinger2021womd,
  title={Large Scale Interactive Motion Forecasting for Autonomous Driving: The {Waymo Open Motion Dataset}},
  author={Ettinger, Scott and Cheng, Shuyang and Caine, Benjamin and Liu, Chenxi and Zhao, Hang and Pradhan, Sabeek and Chai, Yuning and Sapp, Ben and Qi, Charles R. and Zhou, Yin and others},
  booktitle=ICCV,
  pages={9710--9719},
  year={2021}
}

@article{treiber2000idm,
  title={Congested Traffic States in Empirical Observations and Microscopic Simulations},
  author={Treiber, Martin and Hennecke, Ansgar and Helbing, Dirk},
  journal={Physical Review E},
  volume={62},
  number={2},
  pages={1805--1824},
  year={2000}
}

@article{yang2026posttraining,
  title={Post-Training in End-to-End Autonomous Driving},
  author={Yang, Ruining and Wang, Muxing and Chen, Yixiao and Guo, Tongfei and Xu, Yi and Cui, Can and Yang, Zichong and Zhang, Yitian and Wang, Ziran and Fu, Yun and Su, Lili},
  journal={arXiv preprint arXiv:2607.08072},
  year={2026}
}

@inproceedings{peng2024improving,
  title={Improving Agent Behaviors with {RL} Fine-Tuning for Autonomous Driving},
  author={Peng, Zhenghao and Luo, Wenjie and Lu, Yiren and Shen, Tianyi and Gulino, Cole and Seff, Ari and Fu, Justin},
  booktitle=ECCV,
  pages={165--181},
  year={2024}
}

@inproceedings{wei2026parkdiffusionpp,
  title={{ParkDiffusion++}: Ego Intention Conditioned Joint Multi-Agent Trajectory Prediction for Automated Parking using Diffusion Models},
  author={Wei, Jiarong and Rehr, Anna and Feist, Christian and Valada, Abhinav},
  booktitle=ICRA,
  year={2026}
}

@article{wei2026copark,
  title={{CoPark}: Learning Reactive Parking via Self-Play},
  author={Wei, Jiarong and Chen, Yanxing and Song, Sinuo and Wu, Yin and Rehr, Anna and Valada, Abhinav},
  journal={arXiv preprint arXiv:2606.04149},
  year={2026}
}

@article{sisask2026selfplay,
  title = {What Emerges and What Breaks in Self-Play Driving},
  author = {Sisask, Laur and Tampuu, Ardi and Matiisen, Tambet},
  journal = {arXiv preprint arXiv:2608.30819},
  year = {2026},
}

@article{kumar2019rcp,
  title={Reward-Conditioned Policies},
  author={Kumar, Aviral and Peng, Xue Bin and Levine, Sergey},
  journal={arXiv preprint arXiv:1912.13465},
  year={2019}
}

@inproceedings{chen2021dt,
  title={Decision Transformer: Reinforcement Learning via Sequence Modeling},
  author={Chen, Lili and Lu, Kevin and Rajeswaran, Aravind and Lee, Kimin and Grover, Aditya and Laskin, Misha and Abbeel, Pieter and Srinivas, Aravind and Mordatch, Igor},
  booktitle=NEURIPS,
  volume={34},
  pages={15084--15097},
  year={2021}
}

@inproceedings{rowe2024ctrlsim,
  title={{CtRL-Sim}: Reactive and Controllable Driving Agents with Offline Reinforcement Learning},
  author={Rowe, Luke and Girgis, Roger and Gosselin, Anthony and Carrez, Bruno and Golemo, Florian and Heide, Felix and Paull, Liam and Pal, Christopher},
  booktitle=CORL,
  pages={3600--3621},
  year={2024}
}

@inproceedings{anthony2017exit,
  title={Thinking Fast and Slow with Deep Learning and Tree Search},
  author={Anthony, Thomas and Tian, Zheng and Barber, David},
  booktitle=NEURIPS,
  volume={30},
  pages={5360--5370},
  year={2017}
}

@inproceedings{rusu2016distillation,
  title={Policy Distillation},
  author={Rusu, Andrei A. and Colmenarejo, Sergio Gomez and G{\"u}l{\c{c}}ehre, {\c{C}}a{\u{g}}lar and Desjardins, Guillaume and Kirkpatrick, James and Pascanu, Razvan and Mnih, Volodymyr and Kavukcuoglu, Koray and Hadsell, Raia},
  booktitle=ICLR,
  year={2016}
}

@article{koprulu2026cl4ad,
  title={Scaling Curriculum Learning for Autonomous Driving},
  author={Koprulu, Cevahir and Paz, David and Tao, Feng and Guo, Yuliang and Huang, Xinyu and Topcu, Ufuk and Ren, Liu},
  journal={arXiv preprint arXiv:2608.22549},
  year={2026}
}

@inproceedings{yu2025dapo,
  title={{DAPO}: An Open-Source {LLM} Reinforcement Learning System at Scale},
  author={Yu, Qiying and Zhang, Zheng and Zhu, Ruofei and others},
  booktitle=NEURIPS,
  pages={113222--113244},
  year={2025}
}

@article{nie2026awm,
  title={World Models as Adversaries: Multi-Agent Self-Play Fine-Tuning for Robust Motion Planning},
  author={Nie, Tong and Mei, Yuewen and He, Junlin and Tang, Yihong and Sun, Jian and Ma, Wei},
  journal={arXiv preprint arXiv:2607.10630},
  year={2026}
}

@article{nie2026adv0,
  title={{ADV-0}: Closed-Loop Min-Max Adversarial Training for Long-Tail Robustness in Autonomous Driving},
  author={Nie, Tong and Tang, Yihong and He, Junlin and Mei, Yuewen and Sun, Jie and Sun, Lijun and Ma, Wei and Sun, Jian},
  journal={arXiv preprint arXiv:2603.15221},
  year={2026}
}

@article{li2026datacentric,
  title={Data-Centric Autonomous Driving: A Data-Utilization Framework for Learning from Large-Scale Driving Logs},
  author={Li, Yu and Kang, Wenhao},
  journal={Electronics},
  volume={15},
  number={16},
  pages={3517},
  year={2026}
}

@article{ecoffet2021goexplore,
  title={First Return, Then Explore},
  author={Ecoffet, Adrien and Huizinga, Joost and Lehman, Joel and Stanley, Kenneth O. and Clune, Jeff},
  journal={Nature},
  volume={590},
  number={7847},
  pages={580--586},
  year={2021}
}

@article{resnick2018backplay,
  title={Backplay: ``Man muss immer umkehren''},
  author={Resnick, Cinjon and Raileanu, Roberta and Kapoor, Sanyam and Peysakhovich, Alexander and Cho, Kyunghyun and Bruna, Joan},
  journal={arXiv preprint arXiv:1807.06919},
  year={2018}
}

@inproceedings{zelikman2022star,
  title={{STaR}: Bootstrapping Reasoning with Reasoning},
  author={Zelikman, Eric and Wu, Yuhuai and Mu, Jesse and Goodman, Noah D.},
  booktitle=NEURIPS,
  volume={35},
  pages={15476--15488},
  year={2022}
}

@inproceedings{wu2024smart,
  title={{SMART}: Scalable Multi-agent Real-time Motion Generation via Next-token Prediction},
  author={Wu, Wei and Feng, Xiaoxin and Gao, Ziyan and Kan, Yuheng},
  booktitle=NEURIPS,
  volume={37},
  year={2024}
}
